\documentclass[11pt,a4paper,logo]{googledeepmind}

\ifdefined\pdfinfoomitdate
\fi
\ifdefined\pdftrailerid
  \pdftrailerid{redacted}
\fi

\usepackage{subcaption}
\usepackage{placeins}
\usepackage{float}
\usepackage{mathtools}
\usepackage[textsize=tiny]{todonotes}
\definecolor{bagelorange}{HTML}{AE3E06}

\usepackage{arxiv_style}
\setarxivlogo{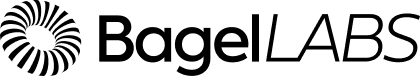}
\setarxivlogowidth{90pt}
\setarxivlogoraise{0pt}
\setarxivdatetext{}

\usepackage[numbers,sort&compress]{natbib}
\hypersetup{plainpages=false}

\theoremstyle{plain}

\theoremstyle{definition}

\theoremstyle{remark}

\title{Paris 2.0: A Decentralized Diffusion Model for Video Generation}
\correspondingauthor{research@bagel.com}

\makeatletter
\renewcommand\AB@authnote[1]{}
\renewcommand\AB@affilnote[1]{}
\makeatother

\author{Ali Rouzbayani}
\author{Bidhan Roy}
\author{Marcos Villagra}
\author{Zhiying Jiang}

\newcommand{\parishfrepolink}{%
  \raisebox{-0.2ex}[0pt][0pt]{\includegraphics[height=1em]{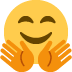}}%
  \,\href{https://huggingface.co/bageldotcom/paris2}{\textcolor{bagelorange}{\texttt{bageldotcom/paris2}}}%
}

\affil{Bagel Labs (bagel.com)}

\makeatletter
\renewcommand{\maketitle}{\bgroup\setlength{\parindent}{0pt}
  \begin{adjustwidth}{0pt}{24pt}
    \begin{flushleft}
      {
        {\raggedright \titlefont \@title\par}%
        \vskip11pt
        {\raggedright\Authfont Ali Rouzbayani, Bidhan Roy, Marcos Villagra and Zhiying Jiang\\[\affilsep]
        \Affilfont Bagel Labs (bagel.com)\quad \parishfrepolink\par}%
        \vskip20pt%
      }%
    \end{flushleft}
  \end{adjustwidth}
  \egroup
  {%
    {\abscontent}
  }%
  \thispagestyle{firststyle}
}%
\makeatother

\begin{abstract}
We present Paris 2.0, the first video generation model pre-trained through decentralized
computation. Its training recipe builds upon Paris 1.0 \citep{jiang2025paris},
the first ever open-weight Decentralized Diffusion Model (DDM), which showed that image
generation can be trained without a monolithic GPU cluster. However, temporally coherent video
generation had remained an open problem under decentralized training, and Paris 2.0 closes it.

In low-resolution text-to-video training, against a monolithic model trained on the same data
under a matched total compute budget, Paris 2.0 cuts Fréchet Video Distance (FVD) from
\textbf{$561.04$} to \textbf{$279.01$}, a
\textbf{$\sim 2.0\times$} improvement, and lifts CLIP text-video similarity and aesthetic score.
\end{abstract}

\begin{document}
\maketitle

\begin{figure}[H]
  \centering
  \begin{minipage}{\linewidth}
    \centering
    \includegraphics[width=\linewidth]{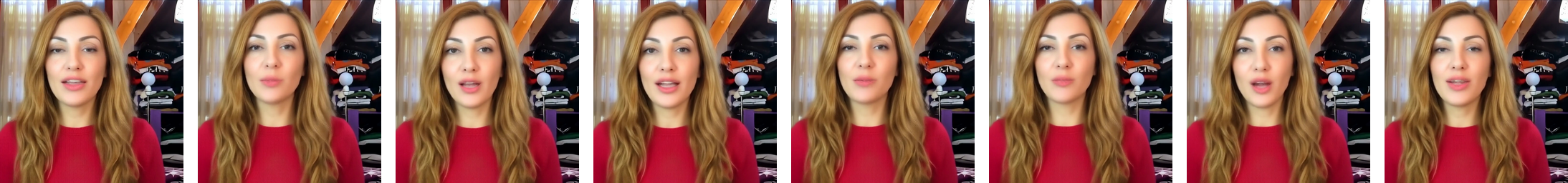}\\[-2pt]
    {\fontsize{4.5}{5.2}\selectfont
    A woman with long, blond, wavy hair is speaking directly to the camera. She is wearing a red
    sweater. While talking, her facial expressions changing as she speaks. The background is a
    cluttered room.}
  \end{minipage}\\[4pt]

  \begin{minipage}{\linewidth}
    \centering
    \includegraphics[width=\linewidth]{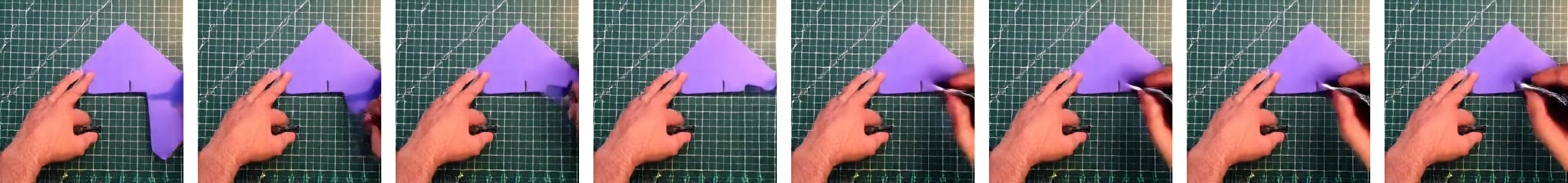}\\[-2pt]
    {\fontsize{4.5}{5.2}\selectfont
    A person's hands performing a paper-folding craft on a green cutting mat with a grid. The
    person uses a black marker to make a small mark on a piece of purple paper that has already
    been folded into a specific shape.}
  \end{minipage}\\[4pt]

  \begin{minipage}{\linewidth}
    \centering
    \includegraphics[width=\linewidth]{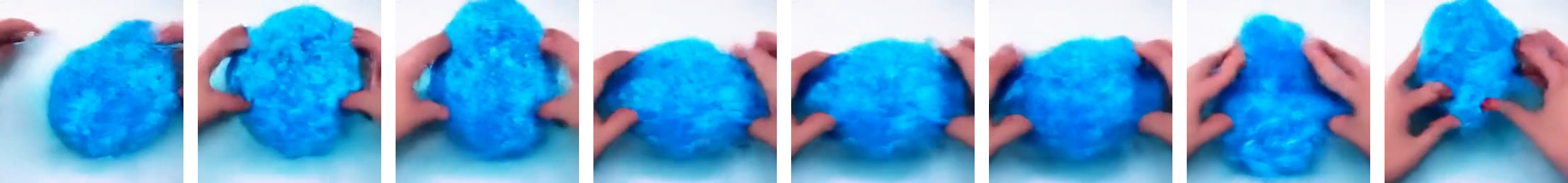}\\[-2pt]
    {\fontsize{4.5}{5.2}\selectfont
    A pair of hands interacting with a translucent, gelatinous slime. The slime is a vibrant
    blue color. The hands are seen stretching, squeezing, and folding the slime,
    demonstrating its gooey and pliable texture.}
  \end{minipage}
  \caption{Qualitative samples from Paris 2.0. Each row shows eight frames from
  one generated video.}
  \label{fig:qualitative-samples}
\end{figure}

\section{Introduction}
\label{sec:introduction}

Video diffusion models are central to media generation, and the same backbones increasingly anchor
the world models that physical AI agents roll forward to predict how their actions reshape a scene. The
prevailing training paradigm nonetheless remains monolithic, in
which a single model is trained on a broad mixture of video datasets and must subsequently generalize
across all prompts at inference. Because the model's parameters are updated at the same time, the
training infrastructure is bound to a homogeneous cluster of supply-constrained GPUs co-located
behind high-bandwidth interconnects.

A Decentralized Diffusion Model (DDM) \citep{mcallister2025ddm} removes that constraint. Paris 1.0
\citep{jiang2025paris} demonstrated the approach for image generation, training an ensemble of
diffusion experts without synchronization among them and routing across them at inference.
However, whether the same recipe could yield temporally coherent video had remained an open
question, since video burdens every expert with motion, longer temporal context, and
substantially heavier latents. Paris 2.0 provides the affirmative answer. We extend the
decoupled-compute training recipe to video and evaluate it directly against the monolithic
alternative it aims to replace.

\begin{figure}[H]
  \centering
  \includegraphics[width=0.95\linewidth]{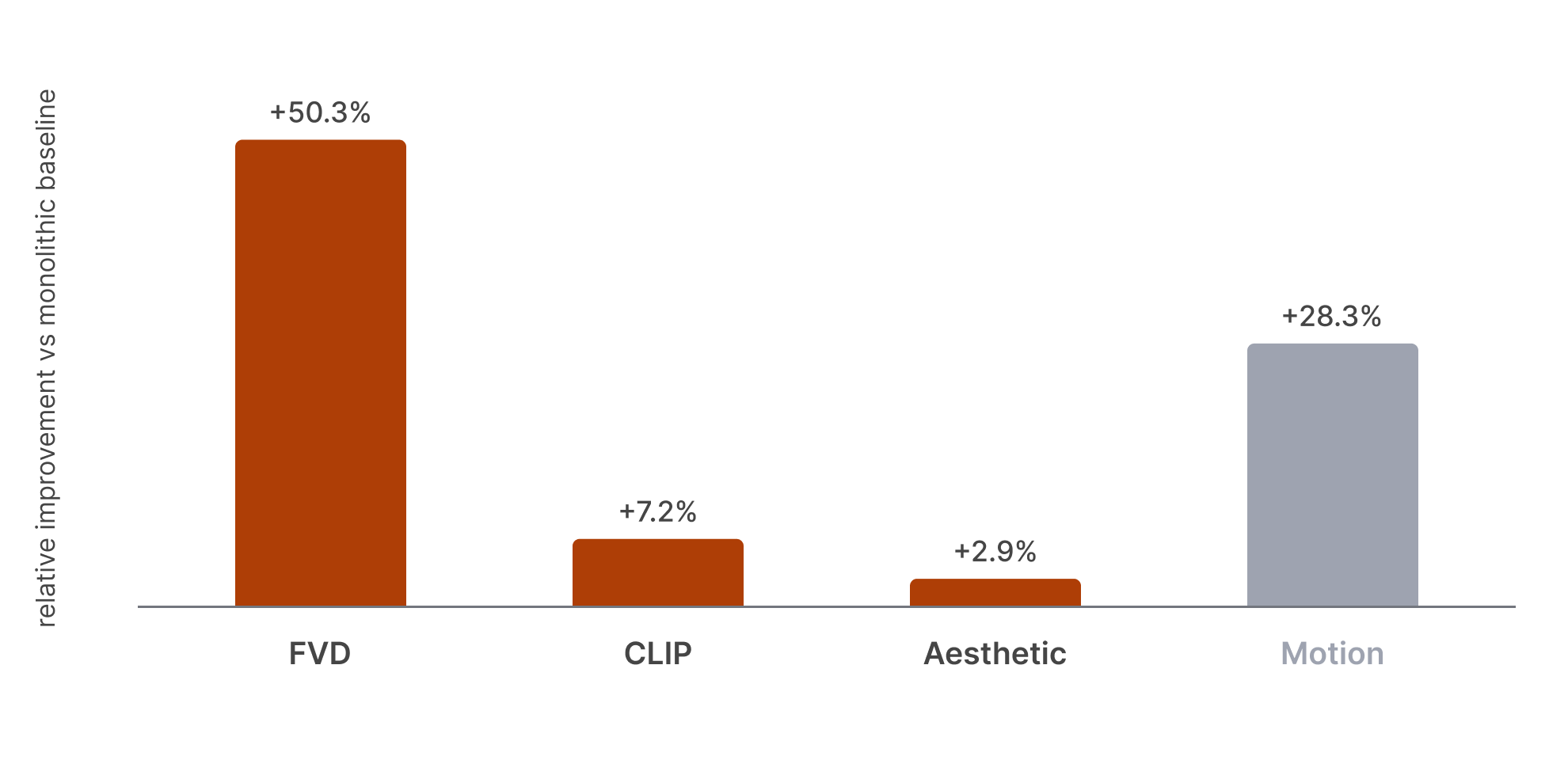}
  \caption{Relative improvement over the monolithic baseline, where a taller bar marks a larger
  gain. Paris 2.0 roughly halves FVD and lifts CLIP text-video and aesthetic scores. Motion is a
  descriptive measure of per-frame displacement magnitude, not of temporal consistency, has no
  preferred direction, and is shown in gray. Absolute values appear in
  Table~\ref{tab:stage1-results}.}
  \label{fig:stage1-bars}
\end{figure}

Under identical data, matched total training compute, and the same generation settings, a Stage 1
three-expert DDM surpasses the monolithic model on major video model benchmarks, as
Figure~\ref{fig:stage1-bars} shows.
Expert capacity trained in isolation does not merely survive the move to video, it outperforms
the single backbone trained on the same data.

\section{Decentralized Diffusion Models}
\label{sec:core-idea}

\begin{figure}[H]
  \centering
  \includegraphics[width=\linewidth]{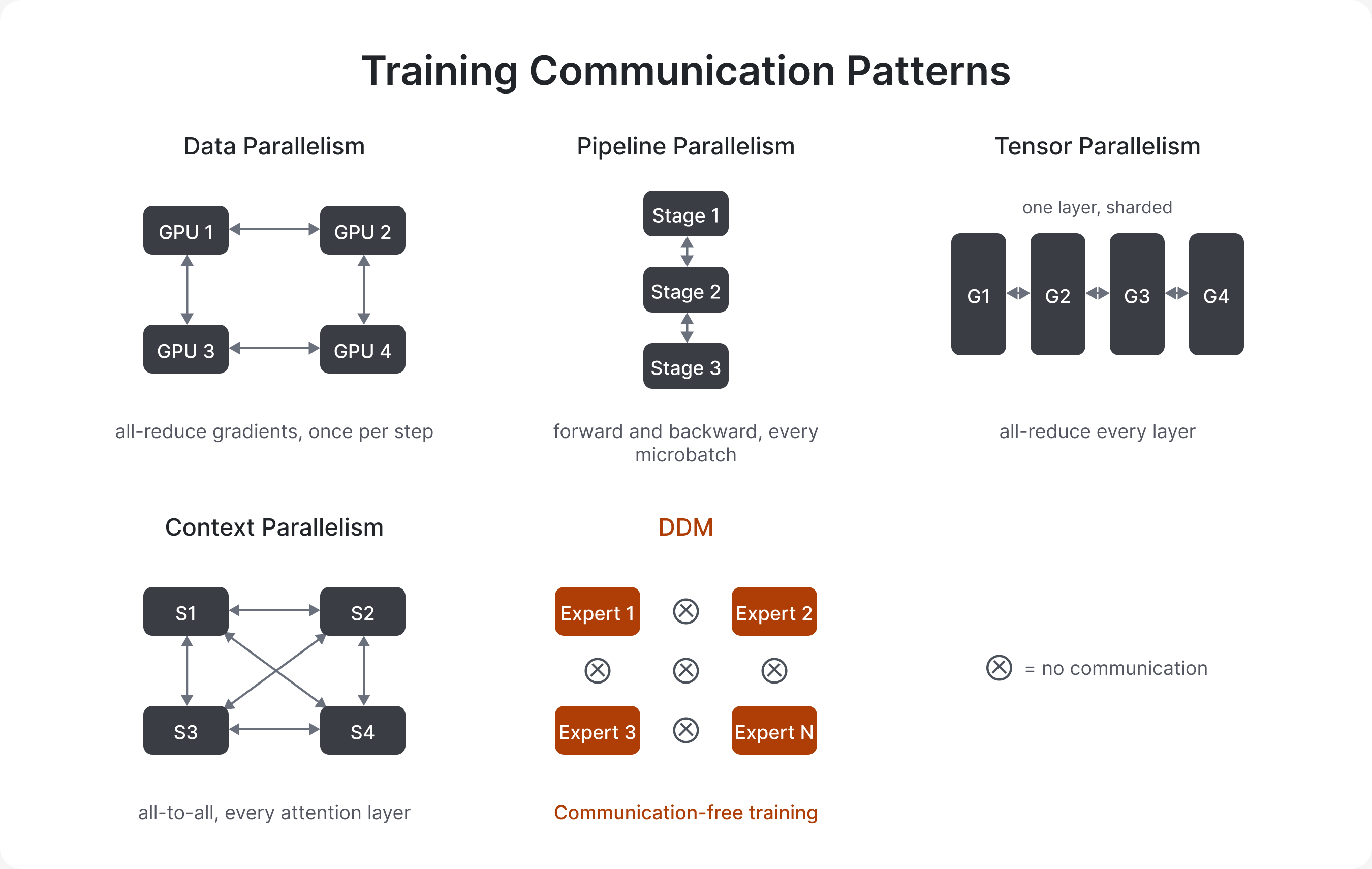}
  \caption{\textbf{Training communication patterns.} Data, pipeline, tensor, and context
  parallelism each synchronize across GPUs during training, from a gradient all-reduce once per
  step to an all-to-all exchange at every attention layer. A DDM requires none of it.}
  \label{fig:comm-patterns}
\end{figure}

A Decentralized Diffusion Model is an ensemble of independent diffusion experts, each a complete
model trained on its own cluster of the data and combined at inference by a learned router. The
experts exchange no gradients, parameters, or activations during training, so each is optimized
in isolation, whereas monolithic training synchronizes across a tightly coupled pool of GPUs at
every step. Eliminating this synchronization lifts the requirement that all compute be
co-located, allowing each expert to train asynchronously on the least costly hardware available,
including spot instances and eligible consumer hardware distributed across clouds and regions, as
Figure~\ref{fig:comm-patterns} illustrates.

During inference, routing is performed at each denoising step, where a lightweight router reads the
current noisy state and selects one or more experts to evaluate the video velocity field. Per-sample
compute stays on the order of a single backbone while total capacity grows with each added expert,
so the training problem becomes horizontally scalable, capacity is added by training another expert
rather than by enlarging one synchronized run. A central and, to our knowledge, novel insight of
Paris 2.0 is that this decomposition is especially well suited to video, where distinct clusters
exhibit different motion patterns, camera behavior, and scene dynamics, so the router can exploit
such specialization without making every sample pay for every expert.

\section{Method}
\label{sec:method}

\subsection{Architecture}
\label{sec:architecture}

Paris 2.0 is built from three pieces, a shared preprocessing stack, an expert pool, and a router.
Videos are encoded into cached causal HunyuanVAE latents \citep{kong2024hunyuanvideo}, where HunyuanVAE is the video
autoencoder that compresses each clip into a compact latent, and prompts are encoded once using
T5-v1.1-XXL and CLIP-ViT-L/14. Expert training consumes these cached tensors directly, so the
expensive perception stack does not sit in the normal training forward path.

Each expert is an $11$B-parameter FLUX-style MM-DiT, the multimodal diffusion transformer
backbone that generates the video, initialized only from FLUX.1-dev image weights
\citep{flux2024} inside an OpenSora-derived training recipe \citep{zheng2024opensora}, and it
operates on packed video latents.

The router is the lightweight counterpart to the expert pool, a DiT-B model of roughly $100$M
parameters. It consumes the current noisy video latent, the diffusion timestep, and the pooled
$768$-dimensional CLIP ViT-L text vector, together with an optional DINOv2 first-frame feature
\citep{oquab2023dinov2}, and produces routing weights over the experts. The experts condition on
the full CLIP and T5-XXL text embeddings, whereas the router reads only the pooled CLIP vector and
never T5. The optional first-frame path, where DINOv2 is a self-supervised vision model that
produces general-purpose image features, lets the same router serve both text-to-video and
image-to-video generation.

\begin{figure}[t]
  \centering
  \includegraphics[width=\linewidth]{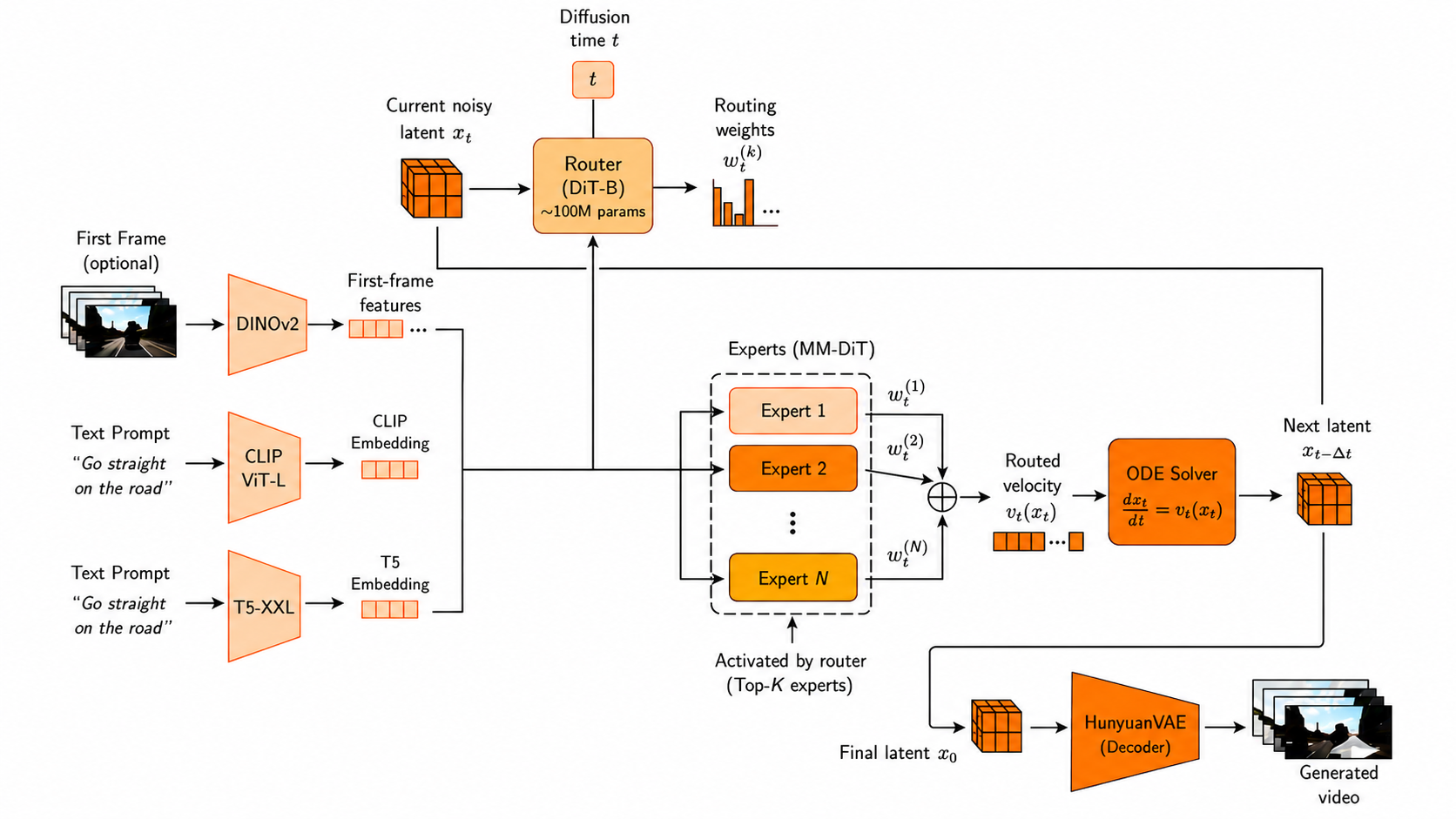}
  \caption{\textbf{The Paris 2.0 inference pipeline.} Text prompts are encoded by T5-XXL and CLIP
  ViT-L, and an optional first frame by DINOv2. At each denoising step the DiT-B router reads the
  current noisy latent, the diffusion timestep, and the pooled CLIP vector, with the optional
  first-frame feature, and produces routing weights that activate the top-K MM-DiT experts. The
  experts condition on the full CLIP and T5-XXL embeddings, while the router uses only the pooled
  CLIP vector. The weighted sum of the experts' velocity fields drives an ODE solver to the next
  latent, and the final latent is decoded to video by HunyuanVAE.}
  \label{fig:architecture}
\end{figure}

\subsection{Training}
\label{sec:training}

Each expert is pre-trained from scratch for video generation on a single data cluster, rather than
fine-tuned from a pre-existing video model, and is optimized with a flow-matching velocity
objective that teaches it to turn noise into video \citep{lipman2023flow}. What matters is not the
exact block schedule or checkpoint format but the training boundary, each expert is confined to
its own cluster.

The router is trained independently of the experts, as a supervised cluster classifier over noisy
latents. From a clean cached latent perturbed with noise at a sampled diffusion timestep under the
training noise schedule, with the paired CLIP text feature, the Stage 1 router predicts the
source-cluster label. This objective is deliberately
simple, and later router stages can use image conditioning and stronger timestep-aware objectives.

\section{Experiments}
\label{sec:experiments}

\subsection{Setup}
\label{sec:setup}

For the Stage 1 study reported here, we use three selected clusters and compare two ways of
using the same data, a monolithic $11$B FLUX-style MM-DiT trained on the union of the clusters,
and three $11$B experts trained separately, one per cluster. Both paths use the same cached
HunyuanVAE/text-embedding pipeline, the same model family, and an aligned flow-matching training
recipe. The comparison is iso-FLOP and iso-data, each expert sees one third of the data with one third of
the monolithic FLOPs, leaving parameter count as the only structural difference between them. Our
prior Bagel Labs analysis of decentralized diffusion at the routing level \citep{villagra2026alignment}
shows that this difference in parameter count is not what drives DDM quality, since full-ensemble
routing across all experts performs strictly worse than sparse routing despite using maximal
capacity.

We evaluate both arms on a deterministic cluster-stratified subset of $N{=}2048$ held-out clips at
$256{\times}256$, using Euler-50 sampling, classifier-free guidance scale $7.5$, and a fixed seed,
so the two models are scored under an identical generation protocol.

\subsection{Results}
\label{sec:results}

\begin{table}[H]
  \caption{\textbf{Low-resolution text-to-video, compute-matched comparison.} Metrics are computed on
  the same $N{=}2048$ cluster-stratified subset. Arrows indicate preferred direction.}
  \label{tab:stage1-results}
  \centering
  \small
  \begin{tabular}{@{}lcc@{}}
    \toprule
    \textbf{Metric} & \textbf{DDM} & \textbf{Monolithic} \\
    \midrule
    FVD $\downarrow$                   & \textbf{279.01}              & $561.04$ \\
    CLIP text-video $\uparrow$         & \textbf{0.2178 $\pm$ 0.0012} & $0.2032 \pm 0.0011$ \\
    Aesthetic $\uparrow$               & \textbf{3.9036 $\pm$ 0.0082} & $3.7950 \pm 0.0077$ \\
    Motion (px/frame)                  & $0.712 \pm 0.057$            & $0.555 \pm 0.043$ \\
    \bottomrule
  \end{tabular}
\end{table}

As Figure~\ref{fig:stage1-bars} plots in relative terms, the DDM cuts FVD, the measure of
distributional realism, from $561.04$ to $279.01$, and lifts CLIP text-video prompt
alignment and aesthetic quality. Motion is a descriptive measure of per-frame displacement
magnitude, not of temporal consistency, and carries no preferred direction. The headline result is
that the
DDM improves distributional and prompt-alignment metrics under the same generation protocol.

\subsection{Ablations}
\label{sec:ablations}

The comparison above tests the DDM against the monolithic baseline. To probe whether the gain is
genuinely about routing across multiple experts rather than ensembling per se, we run two
additional ablations on the same Paris 2.0 expert checkpoints.

\begin{figure}[H]
  \centering
  \includegraphics[width=0.95\linewidth]{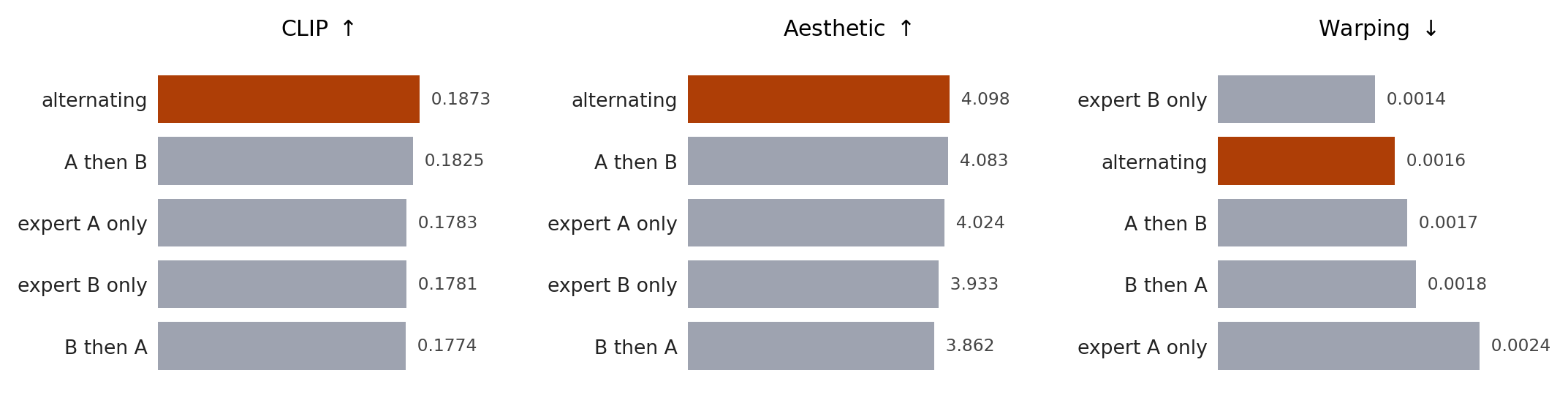}
  \caption{\textbf{Switching schedule ablation.} Manual schedules over two experts with the
  router bypassed, $N{=}40$ video prompts. The alternating schedule wins on all three metrics
  simultaneously.}
  \label{fig:switching-ablation}
\end{figure}

\paragraph{Switching schedule.}
With the router bypassed and manual denoising schedules forced over two experts, an alternating
schedule across denoising steps outperforms either single expert on CLIP, aesthetic, and warping
simultaneously, and $24$ of $40$ prompts prefer a switching schedule over the best single expert,
as Figure \ref{fig:switching-ablation} shows. The asymmetry between expert A then expert B and the
reverse order is large enough to read as directional specialization across denoising time, one
expert prefers high-noise steps and the other prefers low-noise steps. This is the most direct
video-scale evidence that the multi-expert routing principle, not raw parameter count, is what
carries the headline gain.

\paragraph{Expert specialization.}
A second probe asks whether each expert actually learns its assigned data cluster, or whether
it collapses to the marginal distribution. Evaluating one expert checkpoint on its own
cluster's prompts gives CLIP $0.2175$, against $0.1781$ on a generic prompt set on the same
checkpoint, a $0.039$ in- versus out-of-cluster gap. Experts genuinely specialize, which is the
precondition under which a router can extract value at inference.

\section{Discussion}
\label{sec:discussion}

These results show that the DDM recipe survives the move from images to video, where experts
specialized on separate data clusters beat the monolithic backbone, extending our prior result that
the same holds across heterogeneous training objectives \citep{jiang2026heterogeneous}.

The same recipe extends naturally to physical AI. Foundation world models built on video diffusion
backbones must cover far more environments than any single training cluster can hold, and
specializing experts per environment and routing across them at inference fits this regime
directly. Scaling along this axis points toward world models trained on the distributed compute the
world already has rather than the monolithic clusters only a few labs can assemble.

\bibliography{references}
\bibliographystyle{icml2026}

\end{document}